\documentclass[preprint,12pt]{elsarticle}

\usepackage[T1]{fontenc}
\usepackage{graphicx}
\usepackage{subcaption}
\usepackage{algorithm}
\usepackage{algpseudocode}
\usepackage{amsmath}
\usepackage{amssymb}
\usepackage{bm}
\usepackage{hyperref}
\usepackage{array}
\usepackage{xcolor}

\begin{document}

\begin{frontmatter}

\title{Synthetic Data Generation for Anomaly Detection on Table Grapes}

\author[inst1]{Ionut M. Motoi\fnref{fn1}}
\author[inst1]{Valerio Belli\fnref{fn1}}
\author[inst1]{Alberto Carpineto\fnref{fn1}}
\author[inst1]{Daniele Nardi \fnref{fn1}}
\author[inst1]{Thomas A. Ciarfuglia\fnref{fn1}}
\cortext[cor1]{Corresponding author: \href{mailto:ciarfuglia@diag.uniroma1.it}{ciarfuglia@diag.uniroma1.it}}
\affiliation[inst1]{organization={Sapienza University of Rome Department of Computer, Control and Management Engineering
},%Department and Organization
            city={Rome},
            postcode={00185}, 
            state={RM},
            country={Italy}}
\fntext[fn1]{The authors contributed equally to the work.}

%
%
%Abstract
\begin{abstract}
Early detection of illnesses and pest infestations in fruit cultivation is critical for maintaining yield quality and plant health. Computer vision and robotics are increasingly employed for the automatic detection of such issues, particularly using data-driven solutions. However, the rarity of these problems makes acquiring and processing the necessary data to train such algorithms a significant obstacle. One solution to this scarcity is the generation of synthetic high-quality anomalous samples. While numerous methods exist for this task, most require highly trained individuals for setup.

This work addresses the challenge of generating synthetic anomalies in an automatic fashion that requires only an initial collection of normal and anomalous samples from the user—a task that is straightforward for farmers. We demonstrate the approach in the context of table grape cultivation. Specifically, based on the observation that normal berries present relatively smooth surfaces, while defects result in more complex textures, we introduce a Dual-Canny Edge Detection (DCED) filter. This filter emphasizes the additional texture indicative of diseases, pest infestations, or other defects. Using segmentation masks provided by the Segment Anything Model, we then select and seamlessly blend anomalous berries onto normal ones. We show that the proposed dataset augmentation technique improves the accuracy of an anomaly classifier for table grapes and that the approach can be generalized to other fruit types.
\end{abstract}


%%\item Research highlight 1
%%\item Research highlight 2

% Keywords

\end{frontmatter}
\section{Introduction} \label{sec:introduction}
\begin{figure}[ht]
    \centering
    \begin{subfigure}{0.4\textwidth}
        \includegraphics[width=\linewidth]{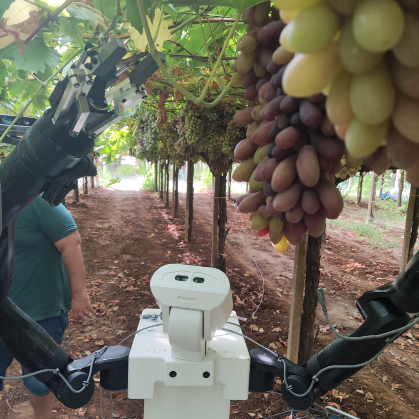}
        \caption{}
    \end{subfigure}
    \begin{subfigure}{0.4\textwidth}
        \includegraphics[width=\linewidth]{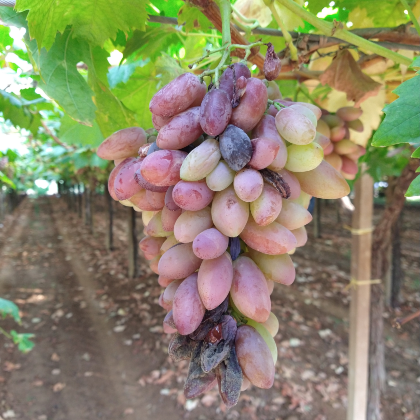}
        \caption{}
    \end{subfigure}
    \caption{Robotic harvesting, as in the EU CANOPIES project, requires high accuracy in detecting anomalous fruits since false negatives could lead to the spread of pest infestations and diseases throughout the orchard. However, given the high variability of possible anomalies and the relative scarcity of naturally occurring examples, synthetic data generation has become an important aspect of addressing this challenge.}
    \label{fig:canopies}
\end{figure}

A crucial decision-support capability in fruit cultivation is the detection and classification of anomalies such as diseases, pest infestations, and other threats (Figure~\ref{fig:canopies}). While the past decade has seen growing interest in applying Computer Vision techniques to agricultural tasks, particularly for labor-intensive processes and decision-support systems \cite{ju2022review,lytridis2021overview,droukas2023survey,ciarfuglia2023weakly}, anomaly detection remains a uniquely challenging problem. Unlike other Computer Vision methods used in Precision Agriculture, such as detection and segmentation \cite{ciarfuglia2023weakly, saraceni2023agrisort, schneider2023skeletonization}, anomaly detection is particularly affected by the issue of data scarcity, a common limitation of data-driven approaches. This challenge is further compounded by the covariate shifts inherent to dynamic, living environments.

% A crucial decision-support capability within fruit cultivation lies in detecting and classifying anomalies like diseases, pest infestations, and similar threats (Figure ~\ref{fig:canopies}). However, while 
% the past decade has witnessed an increasing interest in the application of Computer Vision to agricultural tasks,  
% %\cite{ju2022review,lytridis2021overview}, particularly for labor-intensive tasks \cite{droukas2023survey} and decision support systems \cite{ciarfuglia2023weakly}, 
% %Artificial perception is a key enabler for Robotics applications, and recent advancements have significantly improved perception-related tasks \cite{saraceni2023agrisort,schneider2023skeletonization}. 
% unlike other Computer Vision techniques for Precision Agriculture, such as detection and segmentation  \cite{ciarfuglia2023weakly, saraceni2023agrisort,schneider2023skeletonization}, anomaly detection is even more affected by the data scarcity issue common to data-driven approaches. This challenge is even more pronounced when considering the covariate shifts inherent to growing, living environments.

Data scarcity presents a two-fold challenge in machine learning for agriculture. The first obstacle lies in the initial data collection required to train a machine learning model. The second one is the collection of the data needed to address the ongoing covariate shifts caused by environmental factors, seasonal variations, and other causes \cite{sugiyama2007covariate}. While data scientists can play a crucial role in addressing the first problem, repeatedly relying on their expertise for the second problem becomes less feasible and cost-effective as a long-term solution. Machine Learning Operations (ML-Ops) prioritize developing systems and algorithms capable of adapting to covariate shifts with minimal human intervention \cite{akkem2023smart}.

The detection of anomalous conditions, such as damages, illnesses, pests, and similar, poses an additional difficulty: it is, in general, difficult to find enough data samples of the anomalous condition to be able to train robust detectors in a supervised way, and even more challenging to face the covariate shift problem. For this reason, one of the most common approaches to anomaly detection relies on modeling only the normal condition of the fruit, for which there is plenty of data, in an unsupervised or semi-supervised way and then considering as anomalous all the samples with a distribution significantly different from the training set. In \cite{strothmann2019detection}, an auto-encoder is trained on patches of normal grapevine, and a threshold on the reconstruction loss is extracted from the data to distinguish between the normal and anomalous patches. Combining the patches in the whole image makes it possible to use the loss to define a heatmap that gives some spatial information on the anomaly's location on the fruit. Following this approach, other authors proposed an improvement using Variational Auto-Encoders \cite{miranda2022detection}, enhancing the general performances of the results. On a similar line, \cite{liu2023self} proposes a one-class distribution learning, starting with hyperspectral images and adding a dimensionality reduction step using PCA.

While single-class training makes the data acquisition problem trivial, this approach is not viable for detecting specific anomalies and can lead to many false alarms due to background and illumination noise. For this reason, the more direct approach of training a supervised model to specifically detect the anomalies is still relevant. In \cite{bomer2020automatic}, the authors train a CNN on good and damaged image crops collected at the same resolution and with the same device. However, the collected data for this work is limited and does not show covariate shifts of real applications. Moreover, a self-supervised approach was adopted to address the data collection and labeling issue. On a similar line in \cite{wang2021tomato}, the authors train a YOLO \cite{yolov5} anomaly detector for tomatoes on a web-scraped dataset. While the dataset is quite large for the standards of agricultural applications, it does not model the actual distribution of any field.

Given the aforementioned limitations, many researchers devoted their attention to data augmentation techniques and synthetic data generation. A general augmentation scheme that leverages the often peculiar color differences of anomalous samples has been presented in \cite{choi2022self}. The channel randomization technique is well suited for many fruit anomalies and can be considered complementary to the one we propose. A more traditional approach is followed in \cite{nitin2023developing} and \cite{li2019effective}, where the authors propose augmentation schemes based on various approaches, including rotations, resizing, cropping, and addition of random noise. These approaches can improve the results over the non-augmented case but are not enough when the anomalous samples are really rare and present artifacts due to the simple pasting technique. A different approach is to train generative algorithms to synthesize new samples with a distribution close to the real one. In \cite{noguchi2024mandarin}, the authors apply CycleGAN \cite{zhu2017unpaired} to anomalies in mandarins. While effective, generative synthetic data generation is not trivial to perform, and converging to the right anomalous distribution is not a fully controllable process.

We propose a novel approach that combines classical techniques with foundational models to generate synthetic anomalous samples from real normal and anomalous table grape images collected in the vineyard. The algorithm is designed to operate with minimal manual intervention, requiring only a separation of normal and anomalous training examples—a straightforward labeling task that can be easily performed by farmers or agronomists.  Once provided with this initial dataset, the method is able to produce realistic synthetic anomalies tailored to the field conditions.
The main contributions of this paper are:
\begin{itemize}
\item A novel, texture-focused, semi-automatic algorithm for synthetic data generation
\item A curated dataset of normal and anomalous table grape images collected from the vineyard
\end{itemize}
We validate the proposed method through experiments using a baseline CNN classifier, demonstrating significant improvements in performance metrics, including Balanced Accuracy and F1-score.\footnote{The code and data will be officially released after acceptance and are available upon request}.
\\
\section{Materials and Methods} \label{sec:materials}
\subsection{Experimental Field} \label{sec:experimental_field}
The experimental field consists of two vineyard plots totaling approximately 1.16 hectares in southern Lazio, Italy. The vineyards employ a traditional \textit{Tendone} trellis system with 3x3 m\textsuperscript{2} plant spacing. The vines are mature (over three years old), ensuring full production and representing typical conditions for agronomic tasks like harvesting and pruning. The structures are covered with protective netting, standard in the industry, to safeguard from hail and rain damage. Among the different grape varieties contained in the plots, we focused on the Pizzutello Nero, but the method can be extended to other varieties without loss of generality.
\subsection{Data Collection and Labelling} \label{sec:data_labelling}
In this section, we describe the data collection and labeling process. Images were collected using a smartphone camera during the 2021 growing season. An expert agronomist labeled the initial dataset, focusing on factors such as the presence of visible mold or rot, insect damage, pest infestations, and other quality-impacting issues. From the resulting 88 images of healthy grape bunches and 41 images containing anomalies, we extracted 512x512 pixel patches. The final dataset consisted of 529 "good" grape patches and 166 patches exhibiting anomalies. We employed 3-fold cross-validation for dataset division.
\begin{figure*}[h!]
    \centering
    \includegraphics[width=\textwidth]{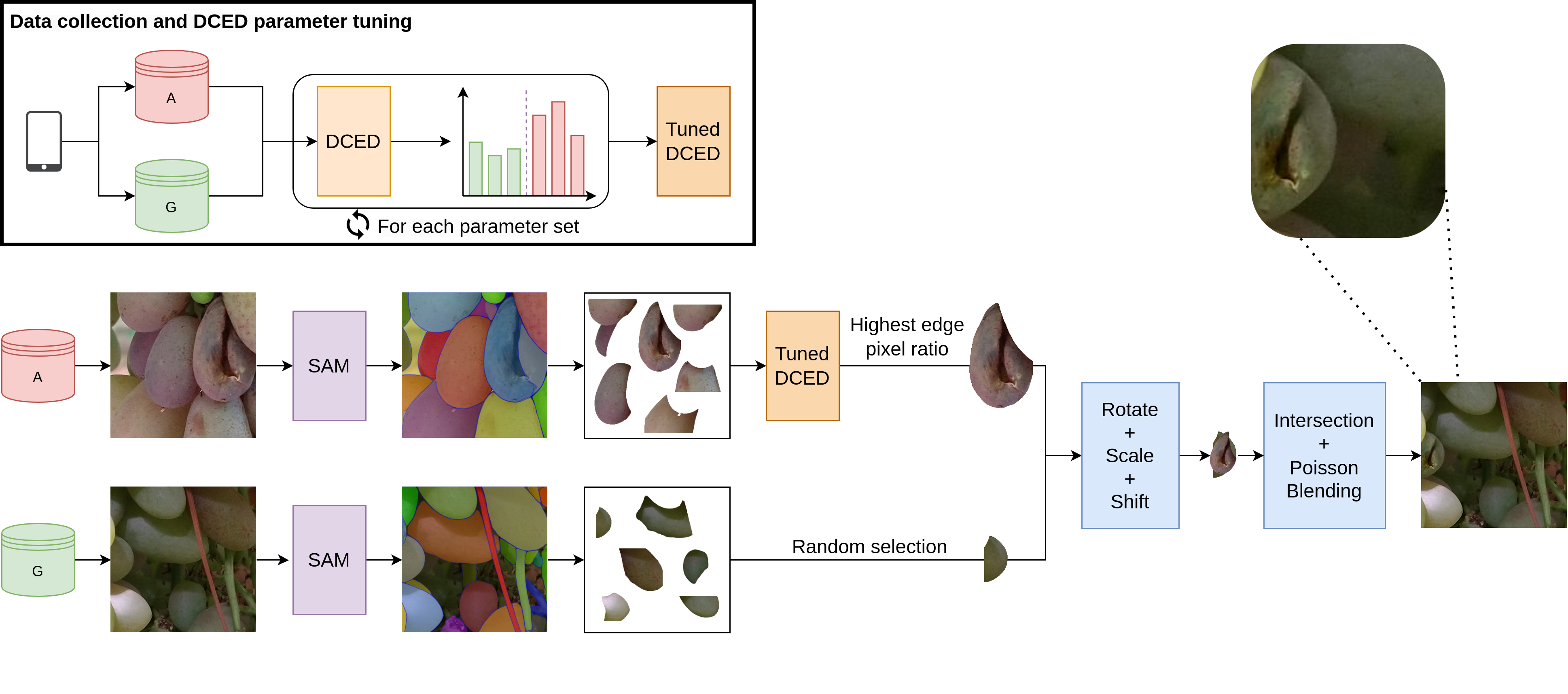}
    \caption{The upper section of the diagram illustrates the preliminary stages of data collection and DCED parameter tuning. The lower section details the synthetic sample generation process. Starting with a pair of real samples from the training set, the system uses SAM to extract their respective masks. The anomalous berry is identified by taking the mask with the highest edge pixel ratio, as determined by the tuned DCED. In contrast, the normal berry is randomly selected. After rotating, scaling, and shifting the anomalous berry, we compute the intersection of the two masks. Finally, we employ Poisson blending to merge the berries and generate a new synthetic sample.}
    \label{fig:system_overview}
\end{figure*}
\subsection{Synthetic Data Generation Procedure}
In this section, we detail the procedure for creating synthetic anomaly data, building upon the real anomaly samples outlined in Section \ref{sec:data_labelling}. We assume that the images are provided by a detector system that isolates and crops the grape bunches within the full images taken on the field. We further assume the ability to extract square patches from these and obtain initial "anomalous" or "normal" labels through inspection by a domain expert. Even with some label noise in the initial dataset, our method is able to generate synthetic samples with a low amount of false positives. Figure \ref{fig:system_overview} visually summarizes the method. In the following sections, we introduce the main components.
\subsubsection{Dual-Canny Edge Detection}
\begin{figure}[h!]
    \centering
    \begin{subfigure}{0.24\textwidth}
        \includegraphics[width=\linewidth]{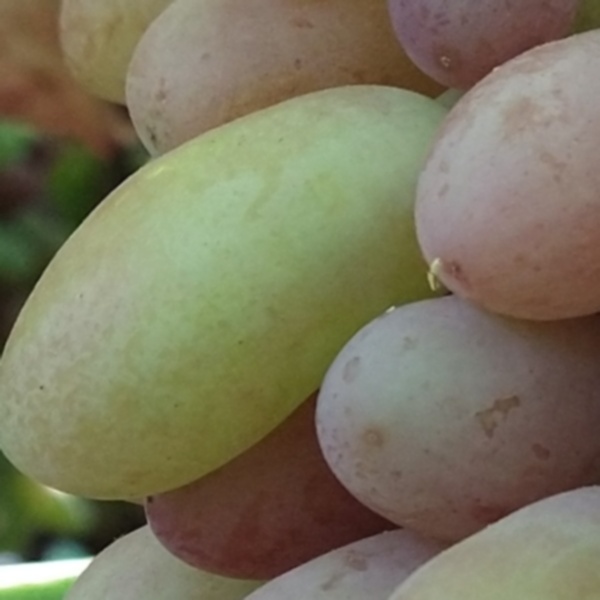}
        \caption{}
    \end{subfigure}
    \begin{subfigure}{0.24\textwidth}
        \includegraphics[width=\linewidth]{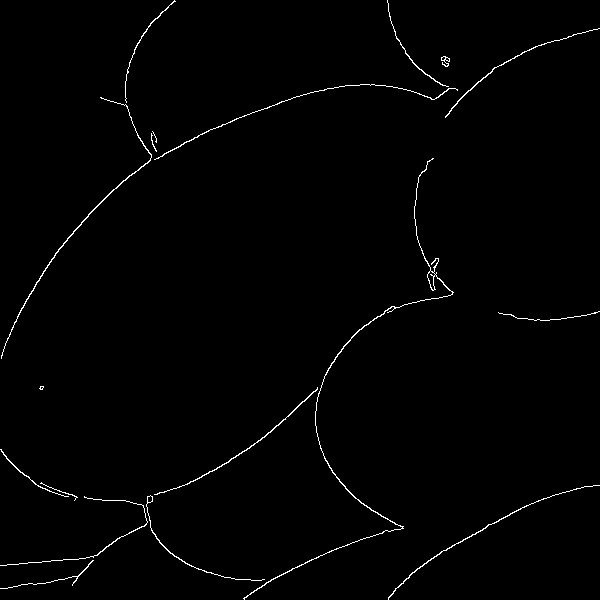}
        \caption{}
    \end{subfigure}
    \begin{subfigure}{0.24\textwidth}
        \includegraphics[width=\linewidth]{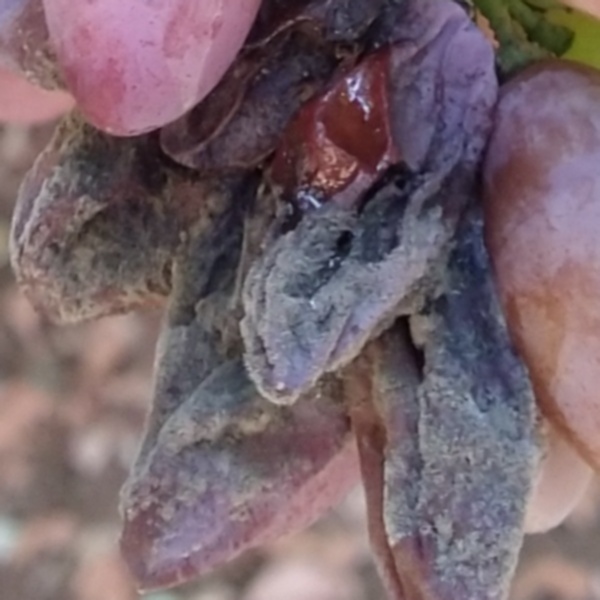}
        \caption{}
    \end{subfigure}
    \begin{subfigure}{0.24\textwidth}
        \includegraphics[width=\linewidth]{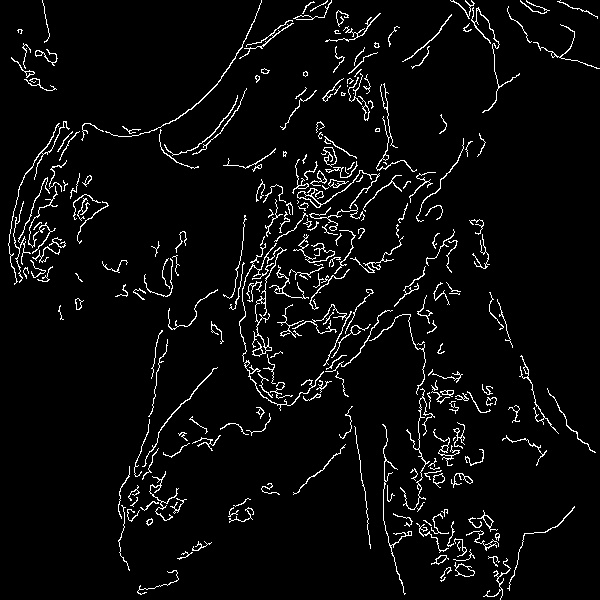}
        \caption{}
    \end{subfigure}\hfill

    \caption{Examples of table grape image patches and relative edge extraction with a canny edge detector. On the left, there is a patch (a) containing berries in good shape and its corresponding edge detection result (b). The edges correspond to the external contour of the berries, while the internal surface is smooth. Image (c) shows a patch containing anomalous berries. The corresponding edge detection result (d) exhibits a more complex edge map, indicating a rougher texture due to defects.}
    \label{fig:Canny_images}
\end{figure}

A key observation for automatically identifying anomalous berries lies in texture: healthy berries tend to have a relatively smooth surface, while anomalous ones are characterized by distinct patterns depending on the specific issue. Therefore, we hypothesized that texture roughness could be measured by edge detection filtering, specifically, Canny Edge Detection (CED) \cite{canny1986edgedetection}. Figure \ref{fig:Canny_images} illustrates the application of CED for this purpose.

CED depends on the choice of some parameters that significantly impact its performance, particularly the hysteresis thresholds, which we'll refer to as $th_{max}$ and $th_{min}$. After Gaussian smoothing with a $K \times K$ kernel for noise reduction, the CED algorithm uses the image gradient to locate potential edges. A double threshold process refines these candidates: pixels exceeding $th_{max}$ are confidently classified as edges, those below $th_{min}$ are discarded, and pixels between the two thresholds are kept only if connected to a confidently classified edge. In this way, edge continuity is ensured despite minor intensity fluctuations.

\begin{figure}[h!]
    \centering
    \begin{subfigure}{0.24\textwidth}
        \includegraphics[width=\linewidth]{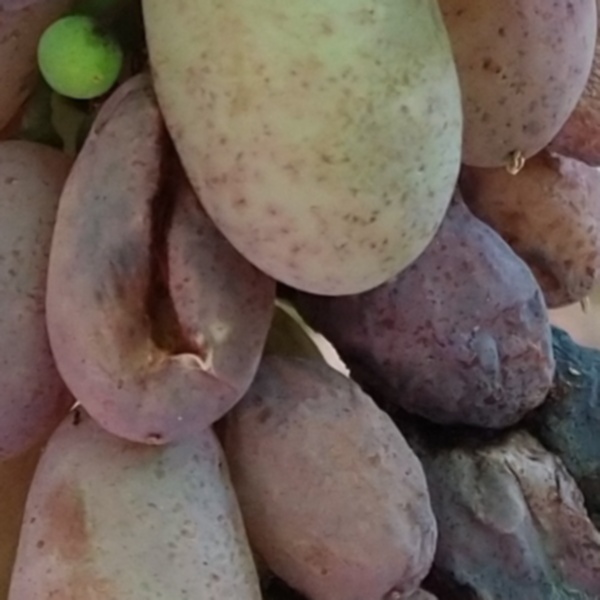}
        \caption{Original}
    \end{subfigure}
    \begin{subfigure}{0.24\textwidth}
        \includegraphics[width=\linewidth]{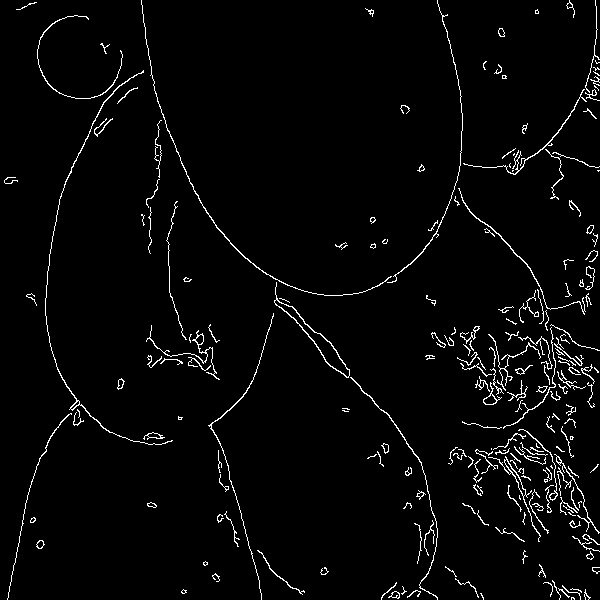}
        \caption{Wide CED}
    \end{subfigure}
    \begin{subfigure}{0.24\textwidth}
        \includegraphics[width=\linewidth]{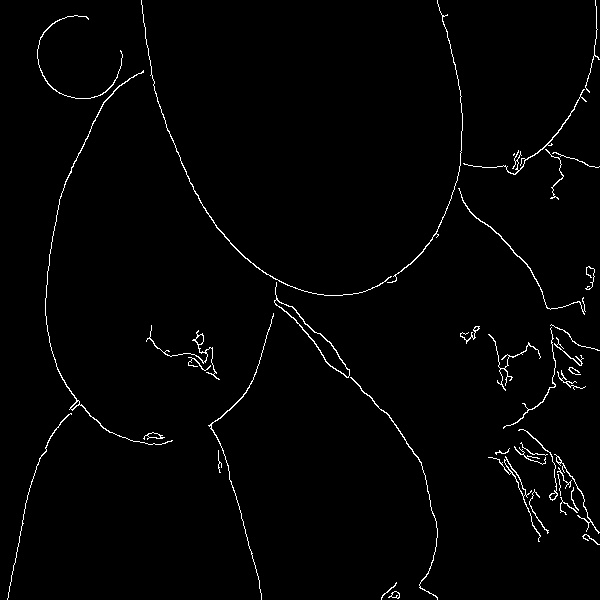}
        \caption{Narrow CED}
    \end{subfigure}
    \begin{subfigure}{0.24\textwidth}
        \includegraphics[width=\linewidth]{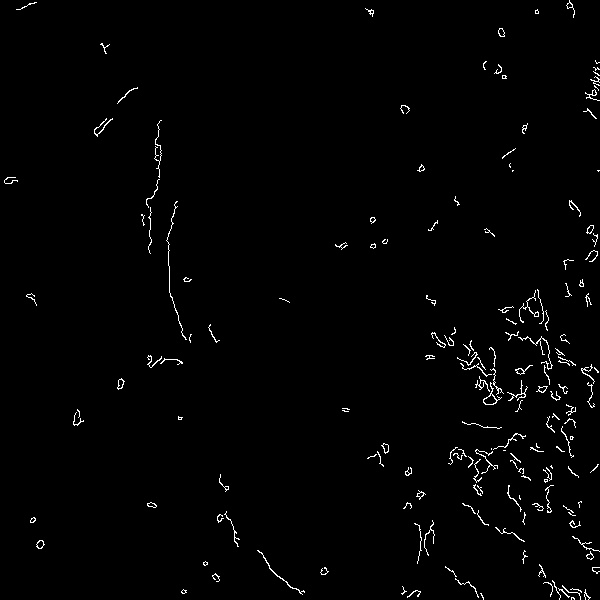}
        \caption{DCED}
    \end{subfigure}\hfill

    \caption{Dual Canny Edge Detection (DCED) example: (a) is the starting anomalous sample, (b) and (c) represent the extracted edges using CED with different thresholds (wide allows for more edges, narrow is more selective), (d) shows the difference between the two edge spaces. It can be seen that, while some border edges are maintained, many edges belong to the anomalous texture.}
    \label{fig:dual_canny}
\end{figure}

While edge-based statistics could be computed from a single CED pass, external berry borders would be present in both normal and anomalous samples. To enhance differentiation, we introduce a Dual-CED (DCED) filter. This involves two CED executions with distinct threshold sets, followed by a subtraction operation. The first pass (wide-CED) employs a wide hysteresis region with lower values for the thresholds $wth_{min}$ and $wth_{max}$. This results in a larger number of edge pixels. The second (narrow-CED) utilizes higher values for the thresholds $nth_{min}$ and $nth_{max}$, resulting in fewer edges. The difference between the two edge maps will mostly contain pixels related to defects in the berries due to the texture observation previously mentioned. Figure \ref{fig:dual_canny} demonstrates the qualitative results of applying the standard CED and DCED.

To optimize DCED parameters for separating "good" and "anomalous" patches, we iterate over different combinations of the kernel size $K$ of the Gaussian blur and of the DCED thresholds ($wth_{min}$, $wth_{max}$, $nth_{min}$, $nth_{max}$). We limited the search space by considering only multiples of 25 in the range [0, 250] and by applying the following constraints: $wth_{max} > wth_{min}$, $nth_{min} \geq wth_{min}$, $nth_{max} > nth_{min}$, and $nth_{max} > wth_{max}$.

For each parameter combination, we analyze edge pixel counts resulting from DCED on the training set and determine the optimal separator using balanced accuracy as our criterion. The combinations of parameters and their corresponding separators are then evaluated on the validation set. Patches whose edge pixel count exceeds the threshold are classified as anomalous and normal otherwise. The final DCED parameter set is selected based on the best overall validation performance and will be used in the synthetic sample generation phase. Additionally, these parameters and optimal edge pixel count threshold serve as a baseline classifier for comparison in Section \ref{sec:experiments}.
\subsubsection{Berry Segmentation and Merging}
\begin{figure}
    \centering
    \includegraphics[width=\columnwidth]{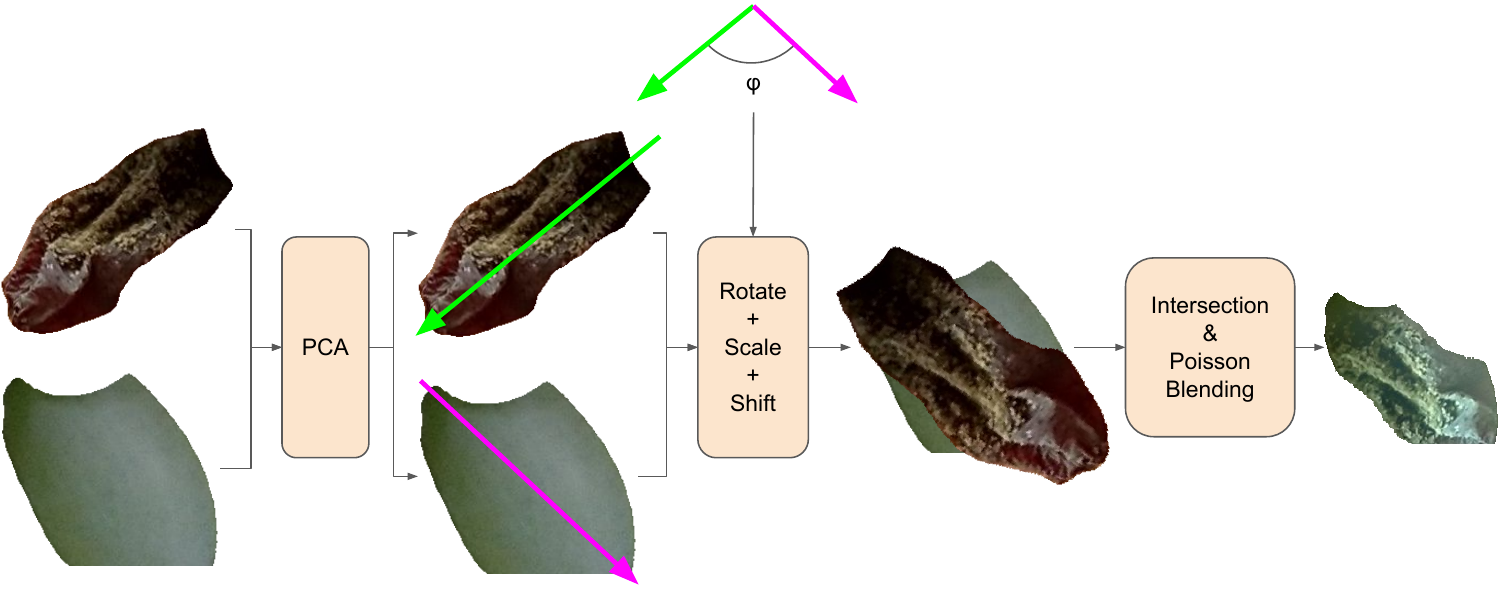}
    \caption{PCA is applied to the two berry masks to determine their primary axis of variation (represented by the green and pink arrows) and compute the angle $\phi$ necessary for aligning the berries. The anomalous grape is then rotated, scaled, and translated to match the normal berry. Finally, the intersection of the two masks is computed, and Poisson blending is employed to seamlessly merge the two images, creating a new anomalous berry.}
    \label{fig:crop-paste-summary}
\end{figure}

The core of our synthetic sample generation technique consists of segmenting damaged berries from anomalous patches and seamlessly pasting them onto berries within good patches, preserving the natural shape and orientation of the destination berry. Figure \ref{fig:crop-paste-summary} outlines this process. 

For our method, we employ the Segment Anything Model (SAM) \cite{kirillov2023segment}, a foundational vision model capable of generating masks for objects within an image. In particular, we use its Automatic Mask Generator mode, which operates without external prompts. We then filter the returned masks, keeping only the top 50\% based on area and only those exceeding the mean area of the masks. {While this filtering approach is straightforward, it proved effective in practice, as most anomalies of interest are comparable in size to the berries. Although background elements may occasionally be included as anomalies, such occurrences are infrequent based on our observations and have a negligible impact on the synthetic data distribution.}

While SAM segments both anomalous and non-anomalous berries (including potentially unwanted elements like branches), we automate the selection of anomalous berries using the DCED with the parameters obtained in the previous phase. Specifically, given the set $\mathcal{S} = \{s_0, \dots, s_n\}$ of $n$ segmented elements extracted from a source patch $\mathcal{P}_S$, we select the mask $s_i$ with the highest ratio of edge pixels.

On the other hand, the destination mask $d_j$ is randomly chosen from the set of $m$ masks $\mathcal{D} = \{d_0, \dots, d_m\}$ extracted by SAM within the healthy patch $\mathcal{P}_D$.

To generate realistic samples containing anomalous berries, we need to seamlessly blend the segmented anomalous source berry onto the normal target berry. While it would be possible to simply paste the anomalous instance at a random position within the destination patch \cite{ghiasi2021simple}, we opt for a more sophisticated approach that accounts for the orientation and size of the berries.

In particular, we determine the longitudinal orientation of the berries by employing Principal Component Analysis (PCA). We apply the PCA algorithm to both the source mask $s_i$ and destination mask $d_j$ to find their principal axes denoted $\mathbf{z}_{s_i}$ and $\mathbf{z}_{d_j}$. The anomalous instance $s_i$ is then rotated by an angle $\phi_{ij}$ and scaled by a factor $\gamma_{ij}$ defined as follows:
\begin{align}
    \gamma_{ij} &= \frac{\mathbf{Area}(d_j)}{\mathbf{Area}(s_i)} \\
    \phi_{ij} &= \arccos{\frac{\mathbf{z}_{s_i} \cdot \mathbf{z}_{d_j}}{|\mathbf{z}_{s_i}||\mathbf{z}_{d_j}|}}
\end{align}

\begin{figure}
    \centering
    \begin{subfigure}{0.25\columnwidth}
        \includegraphics[width=\textwidth]{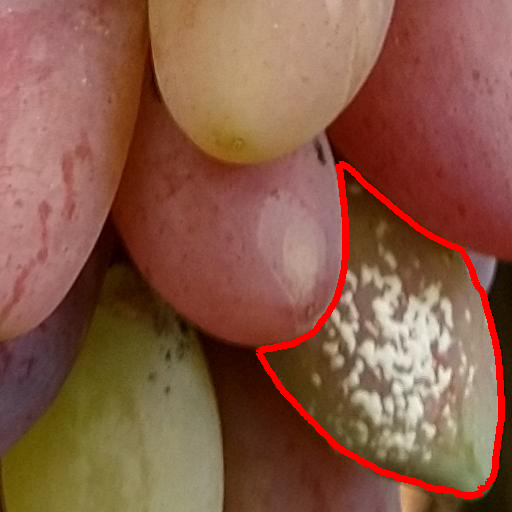}
        \caption{}
    \end{subfigure}
    \begin{subfigure}{0.25\columnwidth}
        \includegraphics[width=\textwidth]{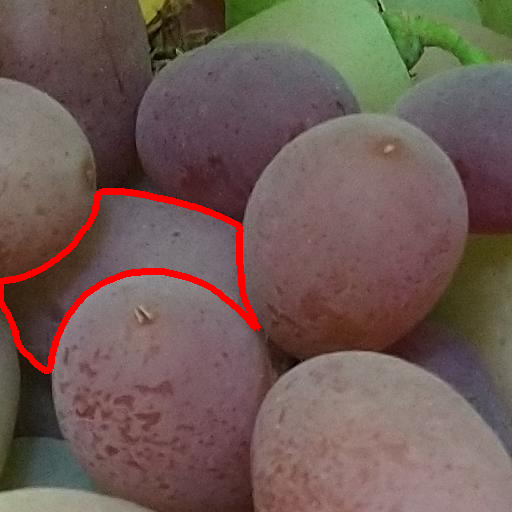}
        \caption{}
    \end{subfigure}
    \begin{subfigure}{0.25\columnwidth}
        \includegraphics[width=\textwidth]{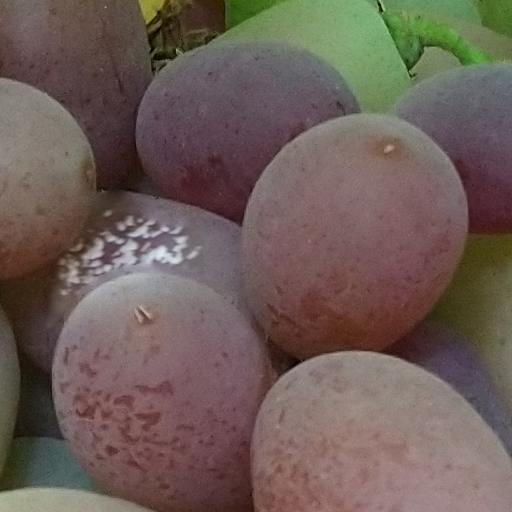}
        \caption{}
    \end{subfigure}
    \begin{subfigure}{0.25\columnwidth}
        \includegraphics[width=\textwidth]{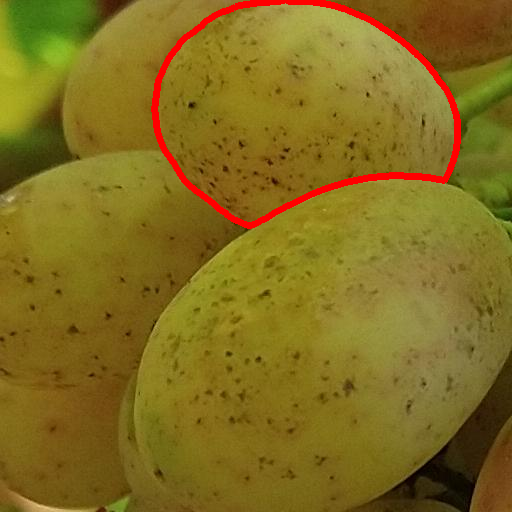}
        \caption{}
    \end{subfigure}
    \begin{subfigure}{0.25\columnwidth}
        \includegraphics[width=\textwidth]{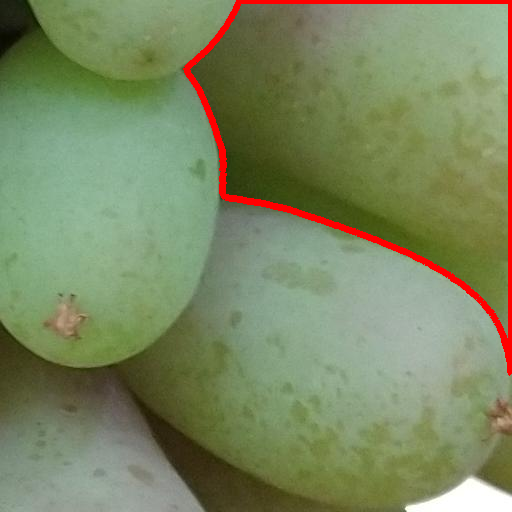}
        \caption{}
    \end{subfigure}
    \begin{subfigure}{0.25\columnwidth}
        \includegraphics[width=\textwidth]{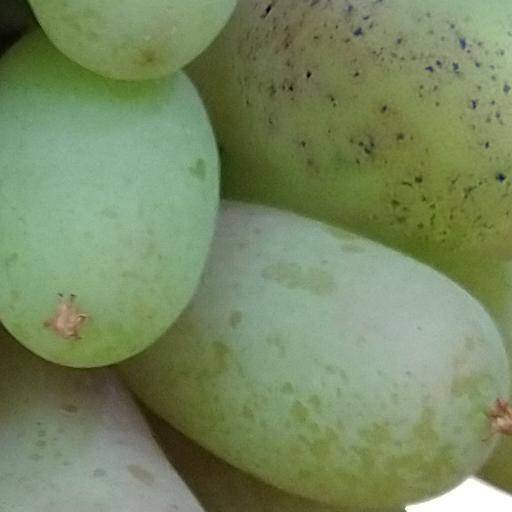}
        \caption{}
    \end{subfigure}
    \begin{subfigure}{0.25\columnwidth}
        \includegraphics[width=\textwidth]{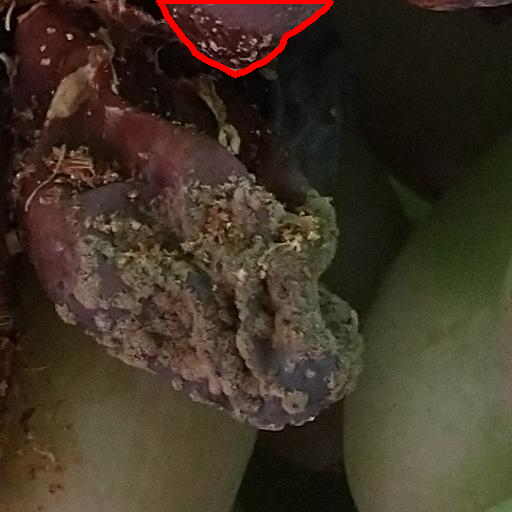}
        \caption{}
    \end{subfigure}
    \begin{subfigure}{0.25\columnwidth}
        \includegraphics[width=\textwidth]{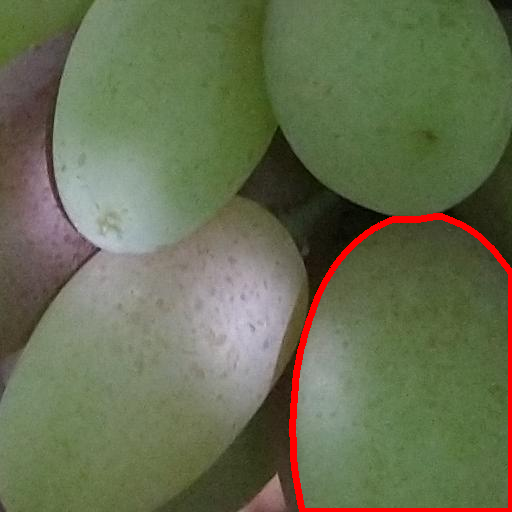}
        \caption{}
    \end{subfigure}
    \begin{subfigure}{0.25\columnwidth}
        \includegraphics[width=\textwidth]{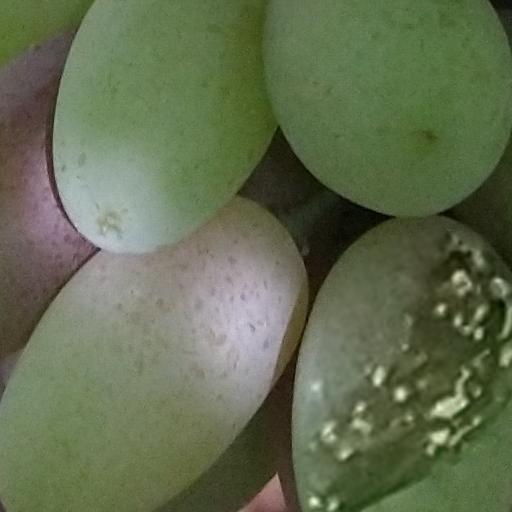}
        \caption{}
    \end{subfigure}
    \begin{subfigure}{0.25\columnwidth}
        \includegraphics[width=\textwidth]{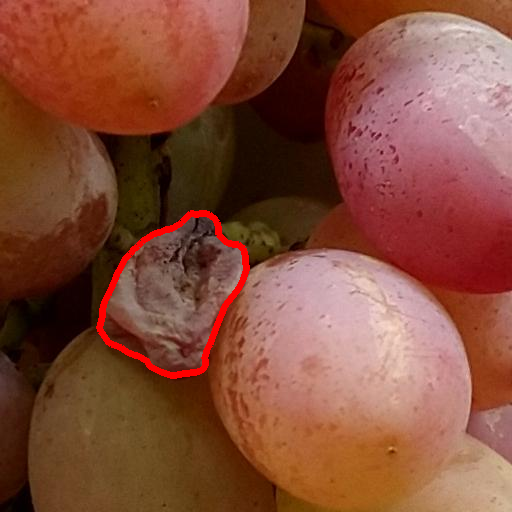}
        \caption{}
    \end{subfigure}
    \begin{subfigure}{0.25\columnwidth}
        \includegraphics[width=\textwidth]{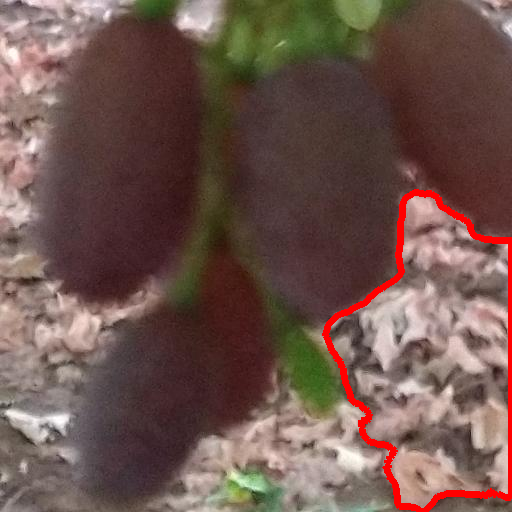}
        \caption{}
    \end{subfigure}
    \begin{subfigure}{0.25\columnwidth}
        \includegraphics[width=\textwidth]{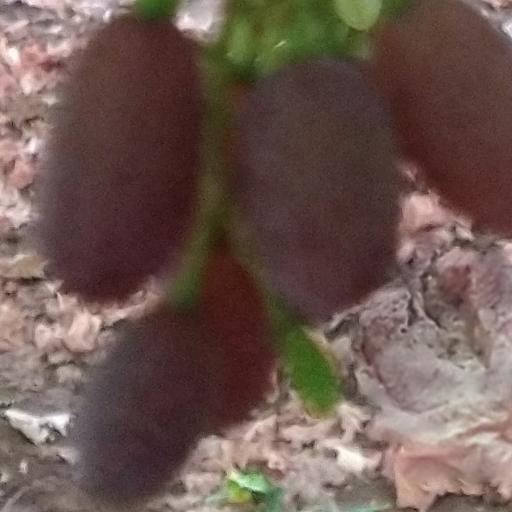}
        \caption{}
    \end{subfigure}
    \caption{Example of the synthetic sample generation. Anomalous berries are seamlessly pasted onto normal berries in target images (b, e, h, k) to create synthetic anomalous images (c, f, i, l). The source and destination masks are highlighted with a red border. (a-c) and (d-f) illustrate successful blending, preserving realistic appearance. (g-i) shows a failure case where the scaling difference between the source and target berries leads to an unrealistic result. (j-l) depicts another failure case where the target area is background instead of a berry, producing a synthetic image that still represents a healthy grape.}
    \label{fig:synthetic_sample}
\end{figure}

We then translate the rotated and scaled anomalous berry, denoted as $s^{\prime}{i}$, on top of the normal berry and take the intersection of their segmentation masks. Finally, to seamlessly merge $s^{\prime}_{i}$ onto $d_j$, a Poisson blending is performed \cite{perez2003poisson}.
%An illustrative example of the generated sample is given in Figure \ref{fig:synthetic_sample}.
Figure \ref{fig:synthetic_sample} provides several illustrative examples of the synthetic sample generation process, showcasing both successful and unsuccessful outcomes. Images (a), (d), (g), and (j) present the original anomalous patches, while (b), (e), (h), and (k) display the target healthy patches with the designated insertion areas outlined in red. The resulting synthetic samples are shown in (c), (f), (i), and (l). Specifically, (a-c) and (d-f) demonstrate successful blending, where the seamlessly integrated anomalous berries retain a realistic appearance within the healthy bunch context. 
%However, (g-i) highlight a failure case where a significant size difference between the source and target berries leads to an unconvincing and unrealistic composite. Another failure scenario is presented in (j-l), where the background is chosen as the target area instead of a healthy berry. This results in a synthetic image that still represents a healthy grape bunch, failing to introduce the intended anomaly. These examples underscore the importance of careful parameter tuning and the limitations of the current approach in handling substantial size discrepancies or background interference.
In contrast, (g-i) illustrate a case where a significant size difference between the source and target berries results in an unconvincing composite, and (j-l) show an example where the background is incorrectly selected as the target area, producing an image that still resembles a healthy grape bunch. These examples highlight the need for improved parameter tuning to better address size mismatches and background interference, which remain key areas for refinement in the current approach.

% \begin{algorithm}
% \caption{Synthetic Data Generation Algorithm} \label{alg:sdg4ad}
% \begin{algorithmic}
% \Require $ N\geq 0$
% \State $max\gets 0$
% \State $best\gets NULL$
% \For{i in Data}
% \State $list\gets SAM(i)$
% \For{j in list}
% \State $j\gets GaussFilter(j,sigma)$
% \State $cw\gets CannyWide(j,dt1,ht1)$
% \State $cn\gets CannyNarrow(j,dt2,ht2)$
% \State $count1\gets CountEdges(cw)$
% \State $count2\gets CountEdges(cn)$
% \State $normdif\gets (count1-count2)/size(j)$
% \If {$max < normdif$} 
%   \State $max\gets normdif$
%   \State $best\gets j$
% \EndIf
% \EndFor
% \State $AddElementToList(DamagedDB, best)$
% \EndFor
% \end{algorithmic}
% \end{algorithm}

\begin{algorithm}[H]
\caption{Synthetic Data Generation}
\begin{algorithmic}[1]
\Require Set of normal images $N$, set of anomalous images $A$, number of synthetic samples per anomaly $n_{syn}$
\State $\text{DCED} \gets \text{TuneDCED}(N, A)$
\State $SyntheticData \gets \emptyset$
\For{$img\_bad \in A$}
    \State $img\_good \gets \text{Sample}(N)$
    \State $bad\_berries \gets \text{filter}(\text{SAM}(img\_bad))$
    \State $bad\_berries \gets \text{SelectEdgiestBerries}(bad\_berries, \text{DCED}, n_{syn})$
    \State $good\_berries \gets \text{filter}(\text{SAM}(img\_good))$
    \For{$i = 1$ to $n_{syn}$}
        \State $good\_berry \gets \text{Sample}(good\_berries)$
        \State $bad\_berry \gets bad\_berries[i]$
        \State $bad\_berry \gets \text{Rotate}(bad\_berry, good\_berry)$ 
        \State $bad\_berry \gets \text{Scale}(bad\_berry, good\_berry)$
        \State $synthetic\_img \gets \text{PoissonBlending}(bad\_berry, good\_berry)$
        \State $SyntheticData \gets SyntheticData \cup synthetic\_img$
    \EndFor
\EndFor
\State \Return $SyntheticData$
\end{algorithmic}
\end{algorithm}

\begin{algorithm}[H]
\caption{SelectEdgiestBerries}
\begin{algorithmic}[1]
\Require masked images of the $berries$, number of berries to select $n$, wide canny edge detector parameters $wide\_params$, narrow canny edge detector parameters $narrow\_params$
\State $edge\_ratios \gets$ Empty list
\For{$berry \in berries$}
    \State $berry \gets \text{GaussianBlur}(berry)$ 
    \State $n\_wide \gets \text{CountNonZero}(\text{CannyEdgeDetector}(wide\_params)$
    \State $n\_narrow \gets \text{CountNonZero}(\text{CannyEdgeDetector}(narrow\_params)$
    \State $n\_berry \gets \text{CountNonZero}(berry)$
    \State $edge\_ratios \gets (n\_wide - n\_narrow) / n\_berry \cup edge\_ratios$
\EndFor
\State  $edge\_ratios \gets \text{SortDescending}(edge\_ratios)$  \
\State \Return $edge\_ratios[:n]$
\end{algorithmic}
\end{algorithm}

\section{Experiments and Results} \label{sec:experiments}
The experiments were conducted using a 3-fold cross-validation over the dataset and we report the mean results over the three experiments.
\subsection{Baselines}
We obtained an initial baseline by applying the DCED filter directly to the image patches. Using the DCED parameters and threshold that performed best on the training split, we applied the filter to the validation split. Table \ref{tab:baselines} reports the results, providing a strong baseline for comparison.

Our primary baseline is a Deep Neural Network (DNN) classifier, obtained by fine-tuning a ResNet18 backbone with its final layer replaced to accommodate the binary classification task. We conducted a hyperparameter sweep to determine the optimal training settings. The model was trained for 100 epochs using the Adam optimizer, with a learning rate of 1e-5 and no weight decay. The learning rate was further reduced to 1e-6 for the final 10 epochs. We used a batch size of 32 and applied random horizontal flips and color jitter for data augmentation. Each model was saved at the epoch with the best balanced accuracy. As shown in Table \ref{tab:baselines}, the ResNet18 baseline achieved significantly better performance, and we will refer to this baseline exclusively in the following sections.

However, it is important to note that in the context of anomaly detection, undetected anomalies can lead to significant costs for farmers (e.g., an undetected illness that spreads). Moreover, improving the performance of an already high-performing classifier is particularly challenging. For these reasons, the use of this baseline is especially relevant.

\begin{table}
    \centering
    \caption{Comparison between the baselines}
    \begin{tabular}{|>{\centering\arraybackslash}p{2.8cm}>{\centering\arraybackslash}p{2.1cm}>{\centering\arraybackslash}p{2.1cm}>{\centering\arraybackslash}p{2.1cm}>{\centering\arraybackslash}p{2.1cm}|} \hline
         &  Balanced Acc.&  F1-Score&  Precision& Recall\\ \hline 
         DCED \& thr. &  76.99 &  60.70 &  48.99 & 80.06\\ \hline
         ResNet18 & 94.99&  92.92&  93.99& 91.88\\ \hline 
    \end{tabular}
    \label{tab:baselines}
\end{table}

\subsection{Experiments}
In this section, we describe the experiments that employed the dataset augmented with synthetic samples. The aim of these experiments is to study the extent to which the addition or substitution of synthetic data generated with the proposed method can improve the performance of the baseline classifier. Given the presence of true anomalous samples in our training set, we aim to measure the degree to which the addition of synthetic data can shift the distribution and to what extent it can be beneficial. For each of the experiments in the 3-fold, we first defined the best parameters for the DCED and used them in the synthetic sample generation process. We generated one synthetic sample for each real anomalous sample. We then randomly sampled 10\%, 25\%, 50\%, and 100\% of the synthetic samples and added them to the original dataset.  In addition, to further examine the covariate shift with the real data, we performed a substitution of 10\%, 25\%, 50\%, and 100\% of the real anomalous data with synthetic data. Together, these two sets of experiments allow us to discuss the effects of the proposed technique of synthetic data generation.

\begin{table}[h]
    \centering
    \caption{Comparison between the baseline and the impact of addition or substitution of synthetic anomaly samples generated by pasting a single berry per sample.}
    % \begin{tabular}{|ccccc|} \hline 
    \begin{tabular}{|>{\centering\arraybackslash}p{2.8cm}>{\centering\arraybackslash}p{2.1cm}>{\centering\arraybackslash}p{2.1cm}>{\centering\arraybackslash}p{2.1cm}>{\centering\arraybackslash}p{2.1cm}|} \hline
         &  Balanced Acc.&  F1-Score&  Precision& Recall\\ \hline 
         ResNet18 Baseline&  94.99&  92.92&  93.99& 91.88\\ \hline 
         Addition 10\%&  \textbf{95.35}&  93.06&  93.36& \textbf{92.78}\\
         Addition 25\%&  95.24&  \textbf{93.30}&  94.48& 92.18\\
         Addition 50\%&  \textbf{95.35}&  93.06&  93.42& 92.77\\
         Addition 100\%&  94.74&  92.40&  93.40& 91.56\\ \hline 
         Substitution 10\%& 72.33& 61.24& \textbf{96.88}& 45.23\\
         Substitution 25\%& 87.69& 84.88& 95.68& 76.53\\
         Substitution 50\%& 91.84& 88.85& 91.93& 86.14\\
         Substitution 100\%& 55.81& 25.93& 56.14& 17.48\\ \hline
    \end{tabular}
    \label{tab:synthetic}
\end{table}

\begin{table}[h]
    \centering
    \caption{Comparison between the baseline and the impact of addition or substitution of synthetic anomaly samples generated by pasting three berries per sample.}
    \begin{tabular}{|>{\centering\arraybackslash}p{2.8cm}>{\centering\arraybackslash}p{2.1cm}>{\centering\arraybackslash}p{2.1cm}>{\centering\arraybackslash}p{2.1cm}>{\centering\arraybackslash}p{2.1cm}|} \hline
         &  Balanced Acc. &F1-Score&  Precision&  Recall\\ \hline 
         ResNet18 Baseline&  94.99&  92.92&  93.99& 91.88\\ \hline 
         Addition 10\%&  95.03&93.27&  95.02&  91.58\\
         Addition 25\%&  \textbf{95.75}&\textbf{93.70}&  94.16&  \textbf{93.39}\\
         Addition 50\%&  95.53&93.62&  94.54&  92.77\\
         Addition 100\%&  95.02&93.25&  95.11&  91.56\\ \hline 
         Substitution 10\%& 73.03&62.74& \textbf{97.36}& 46.43\\
         Substitution 25\%& 87.3&84.22& 94.96& 75.93\\
         Substitution 50\%& 91.82&89.36& 93.87& 85.53\\
         Substitution 100\%& 58.69&32.34& 52.98& 24.76\\ \hline 
    \end{tabular}
    \label{tab:synthetic_3}
\end{table}

All models in this section were trained using the same hyperparameters and augmentations as the baseline. Therefore, we did not specifically optimize them for the augmented dataset. For each experiment, we saved the model at the epoch with the best balanced accuracy on the validation set.

As shown in Table \ref{tab:synthetic}, the average results over the 3-fold cross-validation demonstrate improvements in both balanced accuracy and F1-score with the proposed method. In particular, the maximum gain in terms of balanced accuracy is obtained with 10\% and 50\% of synthetic data, achieving a balanced accuracy of 95.35. Adding 25\% of synthetic data resulted in the best F1-score of 93.30. However, a slight decrease in performance is observed when all of the synthetic samples are added. Therefore, the percentage of synthetic samples added should be tuned to the specific problem and dataset \cite{motoi2024evaluating}, and exploring more advanced algorithms could be beneficial. Conversely, substituting real samples with synthetic ones consistently led to a decrease in performance compared to the baseline. This suggests that while the generated samples aid model training when added to real samples up to a certain limit, they are not sufficient to replace real samples entirely.

% \textcolor{blue}{To reduce the likelihood of mistakenly selecting normal berries instead of anomalous ones during augmentation or selecting a part of the background instead of the healthy berry as the destination of the pasting, which would generate samples of normal grapes labeled as anomalous, we experimented with sampling and pasting three berries for each synthetic sample.} \\
{Since the procedure is fully automatic, there is minimal risk of selecting background elements as the source or target for anomaly generation. While pasting background elements onto good berries can still qualify as a synthetic anomaly, our goal is to avoid scenarios where background is pasted onto background. To mitigate the likelihood of generating images without any valid synthetic anomalies, we experimented with sampling and pasting three berries for each synthetic sample.} \\
 As presented in Table \ref{tab:synthetic_3}, this approach led to slight improvements in both balanced accuracy and F1-score compared to pasting single berries, with the best performance achieved by augmenting the training set with an additional 25\% of synthetic images. However, similar to the single-berry augmentation, substituting real samples with synthetic ones proved detrimental.

\section{Conclusions}
This work addresses the critical challenge of data scarcity in anomaly detection for table grapes by introducing a novel semi-automatic synthetic data generation method. The proposed approach combines the Segment Anything Model with a Dual-Canny Edge Detection filter to segment, identify, and seamlessly blend anomalous berries onto healthy grape images, thereby generating realistic synthetic training samples. Our experiments demonstrated that incorporating these synthetic samples into the training process significantly improved classifier performance, achieving higher balanced accuracy and F1-scores.

Future work could focus on refining the selection process of anomalous and healthy berries, exploring alternative blending techniques, and integrating additional domain-specific knowledge to enhance the quality and diversity of synthetic data. The proposed synthetic data generation method offers a valuable tool for improving anomaly detection in table grape cultivations and can potentially be generalized to other fruit types

\section*{Data Availability Statement}
The corresponding author will provide the data upon reasonable request.

\section*{CRediT authorship contribution statement}
{\bfseries Ionut M. Motoi:} Methodology, Software, Validation, Formal analysis, Investigation, Data Curation, Writing - Original Draft, Writing - Review \& Editing, Visualization {\bfseries Valerio Belli:} Conceptualization, Methodology, Software, Validation, Formal analysis, Investigation, Data Curation, Writing - Original Draft, Writing - Review \& Editing {\bfseries Alberto Carpineto:} Conceptualization, Methodology, Software, Validation, Formal analysis, Investigation, Writing - Original Draft {\bfseries Daniele Nardi:} Funding acquisition, Writing - Review \& Editing {\bfseries Thomas A. Ciarfuglia} Conceptualization, Methodology, Validation, Resources, Writing - Original Draft, Writing - Review \& Editing, Supervision, Project administration, Funding acquisition.

\section*{Declaration of competing interest}
The authors declare that they have no known competing financial interests or personal relationships that could have appeared to influence the work reported in this paper. 

\section*{Acknowledgements}
This work is part of a project that has received funding from the European Union’s Horizon 2020 research and innovation programme under grant agreement No 101016906 – Project CANOPIES.
\\[\baselineskip]
This work has been partially supported by project AGRITECH Spoke 9 - Codice progetto MUR: AGRITECH "National Research Centre for Agricultural Technologies" - CUP CN00000022, of the National Recovery and Resilience Plan (PNRR) financed by the European Union "Next Generation EU".
\\[\baselineskip]
This work has been partially supported by Sapienza University of Rome as part of the work for project \textit{H\&M: Hyperspectral and Multispectral Fruit Sugar Content Estimation for Robot Harvesting Operations in Difficult Environments}, Del. SA n.36/2022.

\bibliographystyle{splncs04} 
 \bibliography{references}
\end{document}